\documentclass[svgnames]{scrartcl}

\usepackage[left=2cm,right=2cm,top=3cm,bottom=3cm]{geometry}
\usepackage{natbib}
\usepackage{graphicx}

\usepackage{scrlayer-scrpage}
\usepackage[
    colorlinks,
    urlcolor=black,
    linkcolor=black,
    citecolor=black,
    filecolor=black,
    pagecolor=black,
    linktocpage=true,
    ]{hyperref}
\usepackage[onehalfspacing]{setspace}

\ihead{J.-M. List}
\ohead{2023}
\ifoot{}
\ofoot{}

\title{Inference of Partial Colexifications from Multilingual Wordlists}
\author{Johann-Mattis List\textsuperscript{12}\\\footnotesize\textsuperscript{1} Chair of Multilingual
Computational Linguistics / University of Passau\\
\footnotesize\textsuperscript{2} Department of Linguistic and Cultural Evolution / Max Planck Institute for Evolutionary Anthropology Leipzig}
\date{2023}
\begin{document}
\maketitle

\begin{abstract}

%%% Leave the Abstract empty if your article does not require one, please see the Summary Table for full details.
\noindent The past years have seen a drastic rise in studies devoted to the investigation of
colexification patterns in individual languages families in particular and the languages of the
world in specific. Specifically computational studies have profited from the fact that
colexification as a scientific construct is easy to operationalize, enabling scholars to infer
colexification patterns for large collections of cross-linguistic data. Studies devoted to partial
colexifications -- colexification patterns that do not involve entire words, but rather various
parts of words--, however, have been rarely conducted so far. This is not surprising, since partial
colexifications are less easy to deal with in computational approaches and may easily suffer from
all kinds of noise resulting from false positive matches. In order to address this problem, this
study proposes new approaches to the handling of partial colexifications by (1) proposing new models
with which partial colexification patterns can be represented, (2) developing new efficient methods
and workflows which help to infer various types of partial colexification patterns from multilingual
wordlists, and (3) illustrating how inferred patterns of partial colexifications
can be computationally analyzed and interactively visualized.

\noindent\textbf{Keywords:} partial colexification, loose colexification, colexification networks, computational comparative linguistics %All article types: you may provide up to 8 keywords; at least 5 are mandatory.
\end{abstract}

\section{Introduction}

The past years have seen a drastic rise in studies devoted to the investigation of colexification
patterns in individual languages families and the languages of the world.  The concept of
\emph{colexification} has proven specifically useful for computational and quantitative approaches
in lexical typology. The term \emph{colexification} was originally proposed by \citet{Francois2008}
as a cover term for all cases where multiple senses are expressed by one word form, no matter
whether the multitude of senses results from polysemy or homophony. Colexifications can
be easily computed from large collections of lexical data, specifically from multilingual wordlists,
in which a certain number of concepts is translated into several target languages (see
\citealt[22-24]{List2014d}). Through the aggregation of several multilingual wordlists, it is
straightforward to assemble large amounts of cross-linguistic colexification data, as witnessed by
the growth in recent versions of the Database of Cross-Linguistic Colexifications (CLICS,
\citealt{List2018e,Rzymski2020}, \url{https://clics.clld.org}), as well as by the increase in studies
which exploit colexification data assembled from different sources \citep{DiNatale2021,Bao2021}.
Quantitative studies on colexification patterns have also shown that it is straightforward to
extract those colexifications that are most likely to result from polysemy by searching for colexifications recurring across several language families -- as opposed to frequent colexifications inside one and the same language family, wich might reflect wide-spread cases of homophony \citep{List2013a}. This means in turn that large colexification networks can be treated as polysemy networks that give us direct insights into certain aspects of lexical semantics \citep{Youn2016,Jackson2019,Harvill2022}. 

Up to today, however, most studies dealing with colexifications focus on colexifications of
\emph{entire words}. Colexifications involving only certain parts of the words in a given language
-- \emph{partial colexifications}, also called \emph{loose colexifications} \citep{Francois2008} --
have rarely been investigated (see \citealt{Urban2011} for an exception) and rarely been computed
automatically from larger collections of cross-linguistic data (see \citealt{List2022e} for initial
attempts). 

Two major factors seem to account for the problems involving studies with partial or loose colexifications. On the one hand, it is less straightforward to model partial colexifications in networks, since the relations between words that share common parts may at times be asymmetric, with one word being entirely repeated in the other word. Not only are different network types needed to model partial colexification networks, it is also much less straightforward to interpret them. On the other hand, it is difficult to \emph{infer} partial colexifications networks from large collections of cross-linguistic data, since partial commonalities between words easily arise by chance or reflect grammatical distinctions (noun classes, gender marking, part of speech). As a result, a method that naively searches for similarities between words in the same language variety in a large corpus typically provides very densely connected noisy networks in which one barely finds any signal that would be interesting from a semantic or cognitive perspective.
Thus, while it is easy to handle noise due to homophony in the case of full colexification networks by using strict thresholds for the occurrence of particular colexifications in combination with normalized weights, it is difficult to use the same criteria when creating partial colexification networks. 
 
This study attempts to address at least some of these problems by proposing new models with which certain kinds of partial colexification patterns can be represented in networks, and by developing new efficient methods and workflows that help to infer different types of partial colexification patterns from multilingual wordlists. 
Having inferred these patterns, the study further shows how they can be visualized and analyzed.

\section{Materials and Equipment}

%Materials and Equipment (including a list of reagents/ materials and/or equipment required; formulation of any solutions where applicable)

%\begin{itemize}
%    \item we need wordlists
%    \item we need data in cldf
%    \item we need linked wordlists
%\end{itemize}

\subsection{Multilingual Wordlists}

The starting point of our new workflow for the inference of partial colexifications are multilingual
wordlists. A wordlists is hereby understood as a collection of word forms which are arranged by
their meaning. Unlike a dictionary, in which the word form (the \emph{headword}) constitutes the
primary linguistic unit by which data are ordered, a wordlist orders words by their meaning. While a
dictionary starts from the form, following a semasiological or form-based perspective, a wordlist
starts from the meaning, following an onomasiological, or concept-based perspective.  As a result, a
multilingual wordlist allows us to compare how certain \emph{concepts} (which are thought to be
generally comparable across languages, even if this may be problematic in practice) are translated
into certain languages. 
 
The compilation and aggregation of multilingual wordlists has made a remarkable progress during the last decade and the number of digitally available wordlist collections is constantly increasing.
On the one hand, large unified multilingual wordlist collections have been proposed in the past
years \citep{IDS,Dellert2020,WOLD}, on the other hand, standards for cross-linguistic data formats have
been constantly improved \citep{Forkel2018a} and applied to many smaller or growing data collections
\citep{TULED-0.11} and for
the purpose of \emph{retro-standardization} \citep{Geisler2021}.

\subsection{Cross-Linguistic Data Formats}
For the exploration of partial colexification patterns across multiple languages, a modified version of the well-known \emph{Intercontinental Dictionary Series} (IDS) was
prepared \citep{IDS}. While the original version mixes phonetic transcriptions with
language-specific phonological transcriptions and orthographic entries, the entries in the modified
version were semi-automatically converted to the International Phonetic Alphabet in the variant
proposed by the Cross-Linguistic Transcription Systems (CLTS) reference catalog 
(\url{https://clts.clld.org}, \citealt{CLTS-2.3.0}, see \citealt{Anderson2018}). The
conversion was done by applying the Lexibank workflow of creating standardized wordlists in
Cross-Linguistic Data Formats \citep{List2022e}. In this workflow, originally non-standardized
datasets are semi-automatically standardized by applying a mix of software tools (based on
CLDFBench, \citealt{Forkel2020}) and manual
annotation in order to convert the data into the formats recommended by the Cross-Linguistic Data
Formats initiative \citep{Forkel2018a}. 
 
The updated version of the IDS 
provides wordlists for 329 language varieties for up to 1310 concepts. The standardized phonetic
transcriptions consist of a total of 558 distinct sounds (types) which occur 2~902~306 times in the
data (tokens), with an average phoneme inventory size of 50.76 sounds per variety.
Although -- strictly speaking -- partial colexifications could in theory also be identified from
orthographic data, being able to work with a larger multilingual wordlist available in phonetic
transcriptions has two major advantages, even if the transcriptions may contain certain errors. On the
one hand, it is easier to evaluate the findings, if transcriptions are harmonized for one dataset,
on the other hand, knowing that sounds are represented in segmented form makes it easier to select
the thresholds by which partial colexifications are preliminarily accepted or discarded.
The revised version of the Intercontinental Dictionary Series is currently curated on GitHub, where
it can be accessed at \url{https://github.com/intercontinental-dictionary-series/ids-segmented}.
The version used in this study is \texttt{v0.2}
(\url{https://github.com/intercontinental-dictionary-series/idssegmented/tree/0.1}).
 
For developmental purposes and in order to test certain technical aspects of the new methods
proposed here, a smaller wordlist by \citet{Allen2007} was used. This list -- also converted to CLDF
(see \url{https://github.com/lexibank/allenbai}) -- offers data for 9 Bai dialect varieties in
standardized phonetic transcriptions. 

\section{Methods} 

Full colexifications across languages can be handled in an efficient way that has shown to provide
very interesting insights into semantic relations. Partial (or loose) colexifications, however,
suffer from noise, resulting from the fact that partial similarities between words in the same
language may result from a large number of factors (coincidence, grammatical markers) that do not
reflect specific semantic relations between the words in question. As a result, the well-established
workflows for the inference of full colexification networks cannot be used to infer partical
colexification networks. In order to handle this problem, I propose a three-stage approach that
starts from the \emph{modeling} of partial colexifications -- which helps to reduce the search space
and provides a consistent \emph{representation} of distinct types of partial colexifications in
networks --, offers efficient methods for the \emph{inference} of specific partial colexification
types, and finally allows us to \emph{analyse} different kinds of partial colexification networks in
various ways. In this context, modeling, inference, and analysis reflect a general approach to
scientific problem solving in the historical sciences that was inspired by 
its application in evolutionary biology \citep{Dehmer2011x}.

\subsection{Modeling Partial Colexifications Across Languages}

\subsubsection{Major Types of Partial Colexification}

When modeling words as sequences of sounds, we can define major sequence relations in a formal way. Since sequences play a crucial role in many scientific fields -- ranging from computer science \citep{Gusfield1997} via bioinformatics \citep{Durbin2002} to physics \citep{Kruskal1999} -- basic relations between sequences have been independently identified and discussed long ago.  
In the following, we will distinguish the term \emph{partial colexification} from the term \emph{loose colexification} (the latter originally termed by \citealt{Francois2008}). According to this distinction, partial colexifications are restricted to concatenative morphology, while loose colexifications would also allow for non-concatenative (paradigmatic) morphology. 
The narrower notion of partial colexifications has the advantage that we can use existing models and insights from earlier studies on sequence and string relations and adopt them to the notion of partial colexifications.
 
When comparing three fictitious sequences \texttt{ABC}, \texttt{XYABCD}, and \texttt{ZABCEF}, it is easy to see that the first sequence \texttt{ABC} recurs in both the second and the third one. In computer science and bioinformatics, \texttt{ABC} is called a \emph{common substring} of \texttt{XYABCD} and \texttt{ZABCEF}. Since there is no longer substring than \texttt{ABC}, it is furthermore the \emph{longest common substring} between both sequences. Regarding the specific relation between \texttt{XYABCD} and \texttt{ZABCEF}, we can say that they share a common substring of length 3. 
 The sequence \texttt{ABC} also shares substrings of length 3 with the two other sequences \texttt{ZABCEF} and \texttt{XYABCD}. In addition, however, we can see that the sequence \texttt{ABC} is \emph{identical} with the common substring, and we can say that \texttt{ABC} is a part of \texttt{XYABCD} and \texttt{ZABCEF}. While the former relation between sequences (sharing a substring of a given length) is commutable, the latter relation isn't: saying that one sequence A is part of another sequence B is not the same as saying that sequence B is part of sequence A. 

Given their importance for a wide range of scientific and industrial applications, many efficient
algorithms for the computation of common substrings and the identification of part-of relations (or
parthood relations, see \citealt{Hoehndorf2009}) in sequences have been proposed (see the overview
in \citealt[89-121]{Gusfield1997}). 
Common substrings and part-of relations are two fundamental relations between sequences that cover
the notion of partial colexification in parts. The exact relation between common substring and
part-of relations on the one hand and the notion of partial colexification on the other hand depends
on the way in which we define the latter. If we insist that the material shared between words from
the same language should reflect \emph{lexical morphemes} -- that is, smallest form-meaning pairs in
a given language that bear a lexical meaning -- we can say that any instance of a partial
colexification between two word forms in a given language also corresponds to a common substring
relation between the sound sequences representing the two forms. If two sound sequences exhibit a
common substring relation, however, this does not automatically mean that we are dealing with a
partial colexification in this narrower sense, since word forms can have common substrings for other
reasons. The common substring can reflect purely grammatical morphemes (various forms of affixes,
compare verbs in German, sharing all the infinitive ending, like \emph{lauf-en} ``run'',
\emph{geh-en} ``walk'', etc.), or the similarity can be accidental (compare German \emph{Herbst}
``autumn'' sharing a common substring \emph{st} with \emph{Wurst} ``sausage''). 

While it seems that scholars implicitly use the term \emph{loose colexification} in a way that
restricts the relation to shared lexical morphemes across words in the same language
\citep{Francois2008}, it is important to note that such a narrow sense is in fact not needed.
Colexification was deliberately chosen as a term to describe the relation of words with different
senses sharing the same pronunciation. As a result, terms like \emph{partial colexification} or
\emph{loose colexification} should also be used in a neutral, overarching form, while semantically
more interesting relations should be inferred from partial colexification patterns in a second step.

% Hoehndorf2009
% https://bmcbioinformatics.biomedcentral.com/articles/10.1186/1471-2105-10-377

%\begin{itemize}
%    \item present substring colexifications in two flavours: a) substring colexification (word a recurs as a substring in word b) and b) common substring colexification, two words a and b share common substring x
%    \item present the difference with respect to directionality etc.
%    \item discuss that these cases can be efficiently computed with the help of -- among others -- suffix trees (quote the Python library here)
%    \item check for the meaning of "maximal repeat" by Gusfield, since this is an important notion here: two strings are different and share a common substring etc.
%    \item longest maximal pair can be seen as a specific type of colexification here, since we avoid identity (!)
%    \item make a table of possible substring relations and how they overlap
%\end{itemize}

%**Definition** A *maximal pair* in a string `S` is a pair of identical
%        substrings `\alpha` and `\beta` in `S` such that the character
%        to immediate left (right) of `\alpha` is different from the character to
%        the immediate left (right) of `\beta`.  That is, extending `\alpha`
%        and `\beta` in either direction would destroy the equality of the two
%        strings.  A *maximal repeat* `\alpha` is a *substring* of `S` that
%        occurs in a maximal pair in `S`. --- [Gusfield1997]_ §7.12, page 143ff.
% https://github.com/cceh/suffix-tree/blob/master/src/suffix_tree/tree.py

\subsubsection{Affix and Overlap Colexifications}

When trying to develop methods that search for meaningful partial colexifications, that is, partial
colexifications which reflect similar processes of lexical motivation \citep{Koch2001} underlying
the formation of words in a given language, it is useful to work with a narrower notion of shared
similarity than the one reflected in the common substring and part-of relations between sequences
introduced above. 
Thus, instead of searching whether a word form A expressing a concept X is part
of a word form B expressing a concept Y, we can ask if A is a \emph{prefix} or a \emph{suffix} of B,
thus, if A is identical with the beginning or the end of B. Similarly, we can ask if A and B
\emph{overlap}, that is, if they share a common substring which is either a prefix or a suffix of
both sequences.

The search for linguistically relevant affix and overlap colexifications can be further restricted
by setting thresholds for the length of the substring which the sequences colexify, and by setting
thresholds for the length of the remaining parts of the sequences which do \emph{not} colexify.
Here, the fact that our multilingual wordlist is now available in the form of fully segmented,
standardized phonetic transcriptions, comes in handy, since it allows us to set up thresholds for a
certain number of \emph{sounds} rather than a certain number of \emph{symbols} which often reflect
individual sounds only in combination. 

\subsubsection{Representing Partial Colexifications in Networks}
Similar to substring colexifications and part-of colexifications, the fundamental difference between
affix colexification and overlap colexification is that the former entails a \emph{directional
relation} (one sequence is a part of the other sequence, in this specific case, appearing in the
beginning or the end), while the direction of the latter (two sequences share a common part) cannot
be directly determined. When representing affix colexifications in networks, we can account for the
directionality of the relation by using directed network models.
 
In contrast to the well-known undirected weighted network models used for the representation of full
colexification networks \citep{List2013a} which can be easily visualized, both interactively
\citep{Mayer2014} and statically (using edge thickness to account for differences in the weights for
the links connecting individual concepts, see \citealt{List2018e}), weighted \emph{directed}
networks draw a link from one concept \texttt{A} to another concept \texttt{B} only in those cases
where an affix colexification from \texttt{A} to \texttt{B} can be attested. As an example, consider
German the words German \textit{Finger} ``finger'' and \textit{Fingernagel} ``fingernail''. Here,
the former word form is an affix of the latter word form (since the latter word starts with the
former word), and we can therefore draw a link from the concept ``FINGER'' to the concept
``FINGERNAIL'' in an affix colexification network, in which both words in German are attested. When
visualizing these networks, we may have links pointing in both directions (since there might well be
languages in which the word for ``fingernail'' appears as an affix of the word for ``finger'',
although we do not expect to find many examples), and we can use arrows at the tips of the links
indicating the direction of individual links in our network. 
 
\begin{figure}
    \includegraphics[width=\textwidth]{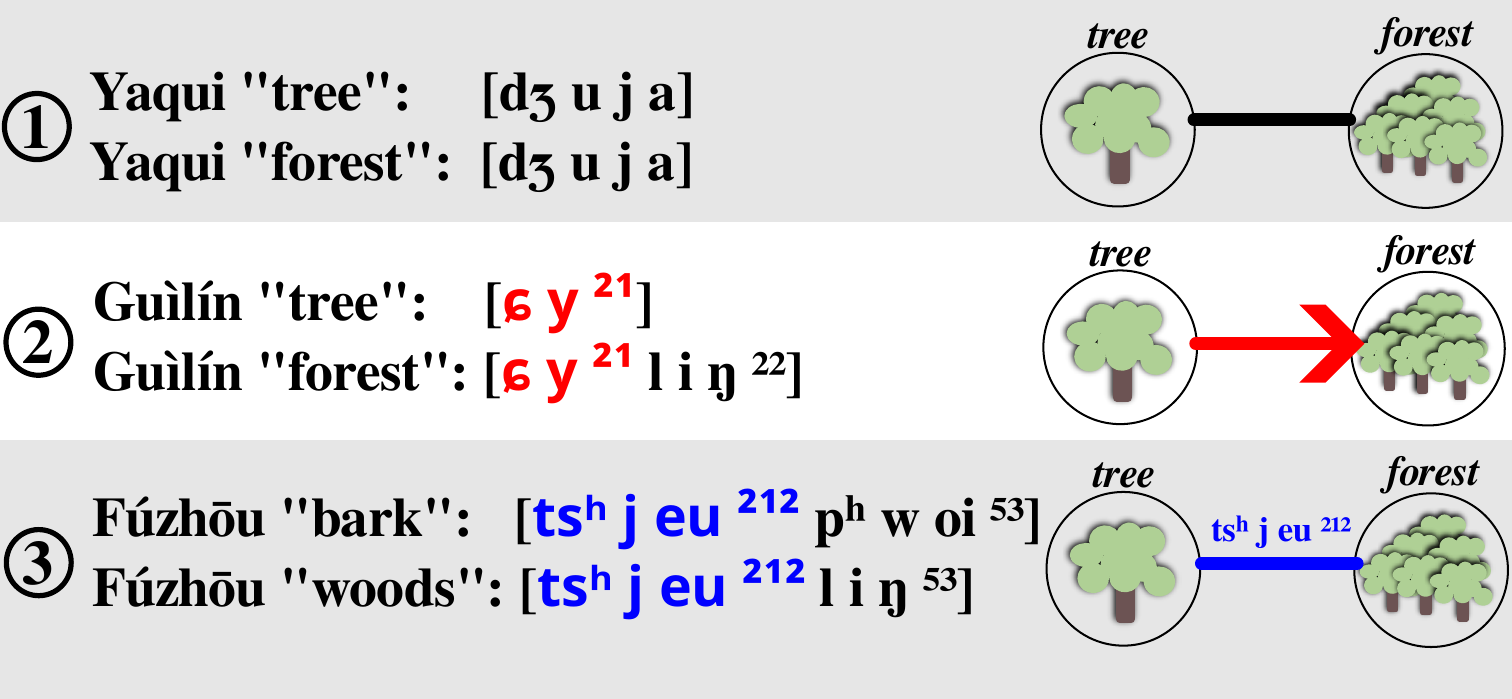}
    \caption{Overview of 3 major colexification type discussed in this stud. (1) provides an example
    for a full colexification in Yaqui (data from CLICS³ \citealt{Rzymski2020}), (2) shows an
    example for the directional representation of affix colexifications with an example from Guìlín
    Chinese (data from \citealt{Liu2007}), and (3) shows an example for overlap colexification in
    Fúzhōu Chinese (data from \citealt{Liu2007}).}\label{Figure1}

\end{figure}

Overlap colexifications can be handled in the same way in which one would handle full
colexifications. The difference is that one should internally store the individual suffixes that
make up for the overlap connection. Thus, while it is enough to store one word form for a
colexification in a full colexification network (since words colexifying two or more concepts are
by definition identical), it is important -- for the sake of transparency -- to indicate the actual
suffix that recurs across two words in an overlap colexification network. Figure \ref{Figure1}
contrasts the three types of colexification networks along with some simplified examples. 
 
\subsection{Efficient Inference of Full and Partial Colexification Networks}

A naive implementation of a simple search for partial colexifications (be they affix or overlap
colexifications) would take all word forms from one language and then compare each word against each
other word in the sample, storing observed commonalities.  While this procedure is easy to
understand and certainly yields the desired results, it is far away from being efficient. As a
result, specifically when dealing with large cross-linguistic data collections, it is advisable to
use efficient search strategies. 
 
\begin{figure}
    \includegraphics[width=\textwidth]{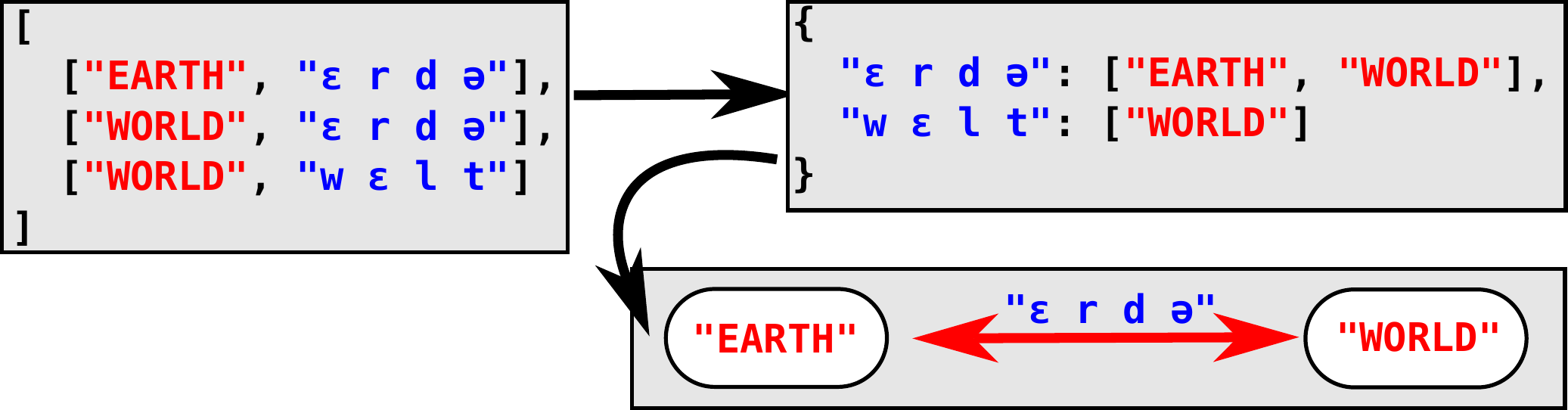}
    \caption{Efficient search for full colexifications using associative arrays.
    Data are represented in JSON format for a wordlist consisting of three German word forms \emph{Erde} ``{EARTH}'', \emph{Erde} ``{WORLD}'', and \emph{Welt}
    ``{WORLD}''.
    The top-left box shows the 
    initial format of the data (a wordlist consisting of two columns, one storing the concept and
    one storing the word form in IPA. The top-right box shows the resulting associative array, in
    which the forms serve as a key and concepts expressing this form are added to the same array as
    a value. The bottom-right box shows the resulting colexification inferred from this example.}
    \label{fig2}
\end{figure}

For the computational of full colexifications, an efficient search strategy consists in the use of
\emph{associative arrays} as a major data structure,  which consists of a key that allows to access
a given value. In the Python implementation used by the CLICS database
\citep{List2018e,Rzymski2020}, the keys consist of the individual word forms for a given language,
while the values are a list of the concepts that the form links to. In order to infer
colexifications for a given language, the method iterates over all words for a given language in a
wordlist and subsequently adds them to the associative array, storing the concept that the word form
expresses in the list that serves as the value. If a given word form has already been added to the
array, the associated list is expanded by adding the new concept in question.  In a second stage,
the method iterates over all keys in the associative array and adds all pairs of diverging concepts
in the list to the growing network of colexifications across several languages.  Detailed
descriptions of this procedure can be found in a tutorial accompanying \citet{Jackson2022} and in
\citet{List2022TBLOG06}. Figure \ref{fig2} shows the structure resulting from applying this method
to a small wordlist of three German words.

In our approach to partial colexifications, we proceed in a similar fashion, by iterating over the
wordlist of each individual language twice. In order to find affix colexifications, however, the
associative arrays are filled with affixes of varying size, and the list serving as the value is
then filled with tuples of the corresponding full word form and its concept. 
The affixes are computed by iterating from the left and the right of the sound sequence representing
the word form. Affix sizes are limited by two thresholds. The first threshold (default set to 2)
limits the minimum size of the affix to 3 sounds. The second threshold makes sure that the size of
the remaining word part is larger than a certain minimum (default set to  2). In combination, both
thresholds guarantee that the affix we infer has a reasonably large size, and that the full word
form to which we link it is also large enough to increase the chances that we detect compounding
structures rather than cases of inflection. With these thresholds, we can detect all potential
\emph{affix candidates} for a given word in a first run and store them in our associative array.
In a second step, we then iterate over all original words in the data, sorted by length, starting
with the longest word. For each of the word forms, we then check if it occurs in the array of affix
candidates. If this is the case, this means that the word appears as the affix of one of word forms
linked as a value to this array, and we can add them to our network, by adding a link from the word
recurring as affix to the word containing the affix. 

\begin{figure}
    \includegraphics[width=\textwidth]{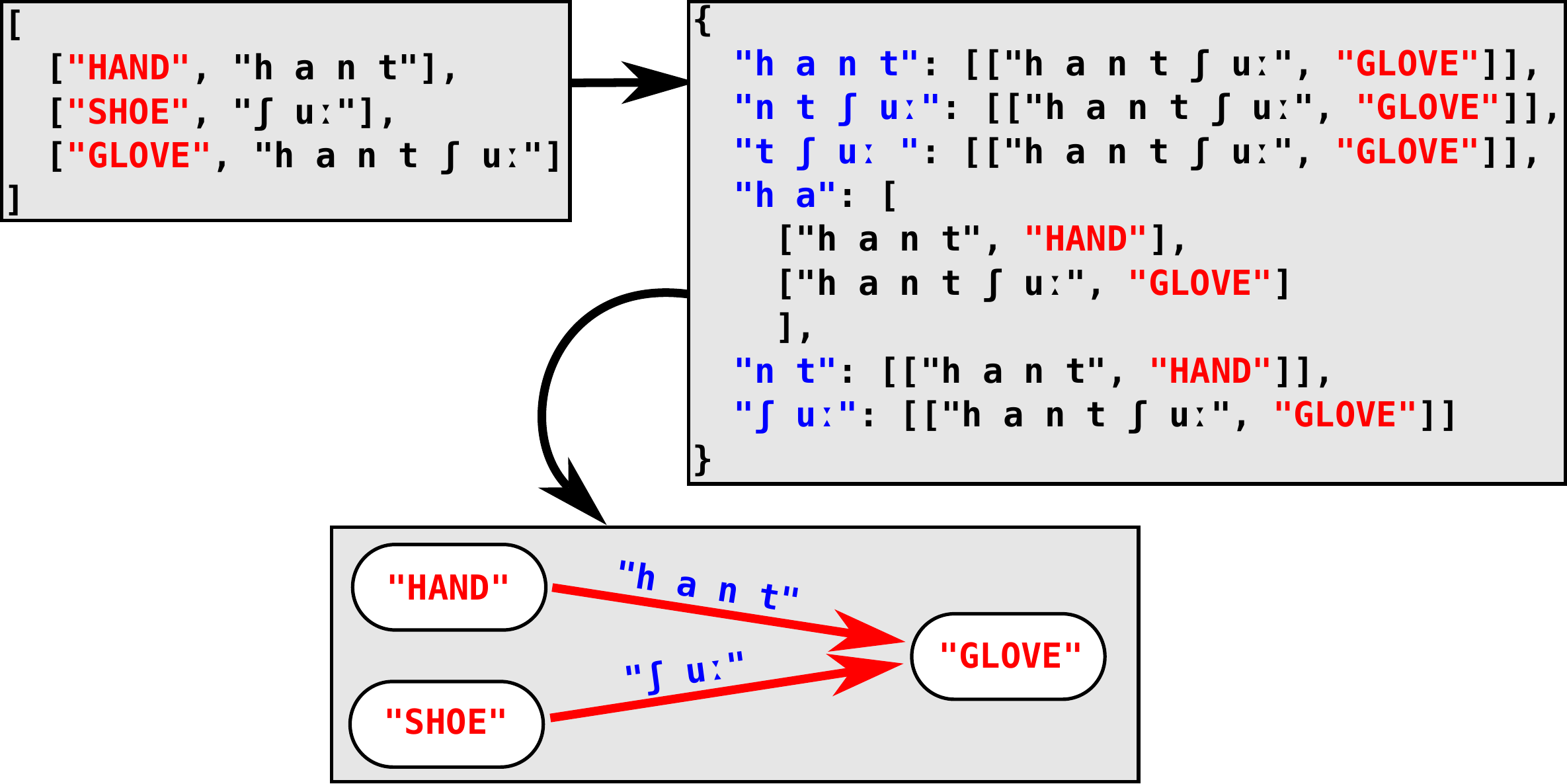}
    \caption{Efficient search for affix colexifications, illustrated for a wordlist of three German
    words \emph{Hand} ``HAND'', \emph{Schuh} ``SHOE'', and \emph{Handschuh} ``{GLOVE}'' (lit.
    ``hand-shoe''). Starting from the original wordlist in the top-left box, each word form is
    represented by all possible prefixes and suffixes that match the two-threshold criterion (see
    text) in the associative array in the top-right box. When iterating over the word forms in the
    original concept list, we find that two words, \emph{Hand} and \emph{Schuh} are stored in the
    array and we can therefore infer an affix relation between the two words and the word
    \emph{Handschuh}, represented in the form of a directed graph in the box at the bottom of the figure.}
    \label{fig3}
\end{figure}

For the computation of overlap colexifications, we pursue the same strategy as for affix
colexifications in the first stage, by populating an associative array with affix candidates for the
word forms in our wordlist. Due to the increase in noise when searching for overlap colexifications
the default thresholds for the length of the affix are set to 4 and the threshold for the length of
the remaining part are set to 3.
In the second stage, we iterate over the array with affix candidates itself, which
has been sorted by the length of the affixes serving as keys in reverse order (starting from the
longest affix found in the data for a given language). 
For each affix, we then compare all word pairs in which this affix recurs and check that neither of
the two forms appears as a suffix or a prefix of the other form. If these conditions are met and the
forms are also not identical (which would correspond to a full colexification), we store the forms
as overlap colexifications along with the affix by which the forms overlap.

As can be easily seen from the descriptions, the complexity of the three methods for the inference
of full colexifications, affix colexifications, and overlap colexifications differs. The search for
full colexifications requires the least amount of computation time, followed by the search for affix
colexifications, and by the search for overlap colexifications. 

With the methods for the inference of full and partial colexifications in individual languages above, we can
construct full and partial colexifications networks by applying the search strategies to multiple
languages and iteratively growing a colexification network, in which we add edges when new edges are
inferred for a particular language, or increase edge weights when edges have been already attested
during the iteration. The networks computed in this form are all annotated in various ways. For the
nodes, we store the number of word forms that can be found in the data, the number of language
families in which these words are attested, and the actual word forms in each language. For the
links between the nodes, we store the number of concrete word forms which exhibit the colexification
relation, the number of language families, in which these colexifications can be found, and the
actual word forms (including the colexifying parts for partial colexifications) in which the
colexifications occur. For affix colexifications, we infer a directed network, while the network for
full and overlap colexifications is undirected.

\subsection{Analyzing Partial Colexification Networks}
In order to understand major differences between full colexification networks and the two new types networks
introduced here, one can compare their \emph{degree distributions}. The degree of a node in a
network is the number of its edges \citep[133-135]{Newman2010}. The weighted degree of a node in a network is the sum of the
edge weights of its edges. While we have only one type of degree for undirected networks, we have
two possible degrees for network with directed edges, the \emph{in-degree} and the
\emph{out-degree}, with the former representing the number (or the sum of the edge weights) of
incoming edges of a given node, and the latter representing the number (or the sum of the edge
weights) of outgoing edges of a given node in the network. In order to compare the degree
distributions of two networks constructed from the same set of nodes, we can compute the Spearman
rank correlation \citep{Spearman1904}, which tell us to which degree those nodes that show a very high degree in one
network also show a high degree in the other network, and to which degree nodes with low degrees in
one network 
also tend to show low degrees in the other one.
 
In addition to the comparison of degree distributions, it is also useful to visualize the networks
and to zoom in to interesting parts that illustrate where major differences can be found. This can
be done quite conveniently now with the help of software packages for network visualization, such as
Gephi \citep{Bastian2009} or Cytoscape \citep{Shannon2003,Smoot2011}. For the visualizations reported here,
Cytoscape is used.

\subsection{Implementation}
The methods reported here are implemented in Python and shared in the form of Python library that
can be used as a plugin to the CL Toolkit package (\url{https://pypi.org/project/cltoolkit},
\citealt{CLToolkit}). CL Toolkit was designed to allow to access CLDF Wordlists that conform to the
standards proposed by the Lexibank repository \citep{List2022e} conveniently from Python scripts or
from the Python interactive console. For the handling of graphs, the NetworkX package was used
\citep{Hagberg2009}, and for the inference of communities, the Igraph package was used
\citep{Csardi2006}. The computation of rank correlations was done with SciPy \citep{SciPy}. 
The supplementary offers access to all data and code necessary
to replicate the results reported here. 

\section{Results}
%There are several ways in which partial colexification networks reconstructed with the methods
%presented here can be analyzed. Following \citet{List2013a}, one could compute \emph{communities},
%i.e., groups of nodes in the network which show more connections among the members of their group
%than with other nodes outside of the group \citep{Newman2004a}. Community structures across
%different colexification networks could then be inspected and compared across the different kinds of
%colexification networks introduced in this study. 

%* we compute colexifications
%* we compute communities for the full colex network
%* we report on computation time

\subsection{Computation Time of Efficient Colexification Inference}
In order to test whether the newly proposed method for the inference of affix
colexifications is indeed more efficient than a conceptually much simpler comparison of all word
against all words in a word list, a small experiment was designed in which the CLDF dataset of Bai
dialects derived from \citet{Allen2007} was analyzed several times and computation times were
calculated. The results of this test indicate that the new method is indeed much more efficient in
terms of computation time than the naive iteration. In various experiments on different Linux
machines, computation time differences show that the naive all-to-all word comparison needs more
than five times as much time than the new efficient approach, while both produce exactly identical
results. While computation time may be less important when working with small datasets
of only about a dozen of languages, it can become a bottleneck when working with large datasets such
as the Intercontinental Dictionary Series. For this reason, the efficient solution proposed here, is
proving very useful. This does not mean, however, that the solution is perfect, and it may well be
the case that there are more efficient solutions available (e.g., using \emph{suffix trees}, see
\citealt[122-180]{Gusfield1997}) that could be implemented in the future.

\subsection{Comparing Degree Distributions}
Having computed colexification networks for full colexifications, affix colexifications, and overlap
colexifications, the Spearman rank correlation was computed for the weighted degree distributions of
all three colexification types, splitting affix colexifications into two types of degree
distributions, the in-degree and the out-degree. The results of this comparison are given in Table
\ref{tab:corr}. As can be seen from this table, two moderate correlations can be observed for the
total of six pairings. The degree distribution of the full colexification network correlates moderately with the
out-degree distribution of the affix colexification network ($r=0.50$, $p<0.0001$), and the degree distribution of
the overlap colexification network correlates moderately with the in-degree distribution of the
affix colexification network ($r=0.42$, $p<0.0001$).

\begin{table}
    \centering
    \begin{tabular}{lllll}
	\textbf{Colexification Type A} & \textbf{Colexification Type B}    & \textbf{Nodes} & \textbf{R} & \textbf{P-Value} \\\hline\hline
Full Colexification                    & Affix Colexification (In-Degree)  & 1308           & 0.0960     & $<$ 0.0001 \\
Full Colexification                    & Affix Colexification (Out-Degree) & 1308           & 0.5034     & $<$ 0.0001 \\
Full Colexification                    & Overlap Colexification            & 1307           & 0.1179     & $<$ 0.0001 \\
Affix Colexification (In-Degree)       & Affix Colexification (Out-Degree) & 1308           & -0.0830    & $<$ 0.0001 \\
Affix Colexification (In-Degree)       & Overlap Colexification            & 1307           & 0.4212     & $<$ 0.0001 \\
Affix Colexification (Out-Degree)      & Overlap Colexification            & 1307           & -0.0488    & 0.0104 \\
\end{tabular}
\caption{Comparing the Spearman rank correlations for the four different kinds of degree
distributions. As can be seen, we can observe significant moderate correlations for Full
Colexifications as compared to the Out-Degree of Affix Colexifications (0.5) and for the In-Degree
of Affix Colexifications compared to Overlap Colexifications (0.42). For the other pairings, no
significant correlations can be observed.} 
\label{tab:corr}
\end{table}

Interpreting these results may not seem straightforward at the first sight. 
The correlation between the weighted degree of concepts in full colexification networks and the
out-degree of concepts in affix colexification networks points to a tendency according to which
concepts that are often fully colexified with other concepts \emph{also} seem to be frequently
\emph{reused} as compounds or affixes in complex words. While this finding may seem to be quite
reasonable or even obvious, it was so far not possible to confirm it in cross-linguistic studies.
Partial colexification networks thus point us to an important property of concepts that tend to
colexify frequently across the languages in the world: their propensity to be reused in word
formation processes to form new words. This property, which I propose to call \emph{lexical root
productivity} (the term is inspired by a discussion with Alexandre François, see
\citealt{List2019PBLOG12} and \citealt{List2019PBLOG11}), plays a key role in lexical motivation, the process underlying the
formation of new word forms in the languages of the world \citep{Koch2001}. 
 
The correlation between
the weighted degree distribution of overlap colexifications with the out-degree distribution of
affix colexifications has an even more straighforward explanation. 
Concepts that exhibit many overlap colexifications across a larger sample of languages are concepts
that are often expressed with the help of compounds or morphologically complex words. 
The same holds for those concepts that have many incoming edges in an affix colexification network.
As a result, the correlation between the two degree distributions is not very surprising. It shows,
however, that both the weighted in-degree of affix colexification networks and the weighted degree
of overlap colexification networks can be used as a proxy to measure the \emph{compoundhood of
concepts} (a term inspired by Martin Haspelmath, p. c.), that is, the tendency of concepts to be expressed by compound words or
morphologically complex words.

\subsection{Inspecting Colexifications through Subgraphs}
While the investigation of the degree distributions already gives us a nice impression about the
commonalities and differences between different kinds of colexification networks, a closer
investigation of smaller parts of the graphs can help us to see these differences much more
clearly. In order to provide a fruitful sample, the Infomap algorithm \citep{Rosvall2008} was used to compute
communities from the full colexification network. In a second step, 23 concepts which show different
properties with respect to their full and partial colexifications, were selected and the
corresponding subgraphs for full, affix, and overlap colexification networks were extracted and
visualized with the help of Cytoscape \citep{Shannon2003,Smoot2011}.  

\begin{figure}
    \includegraphics[width=0.9\textwidth]{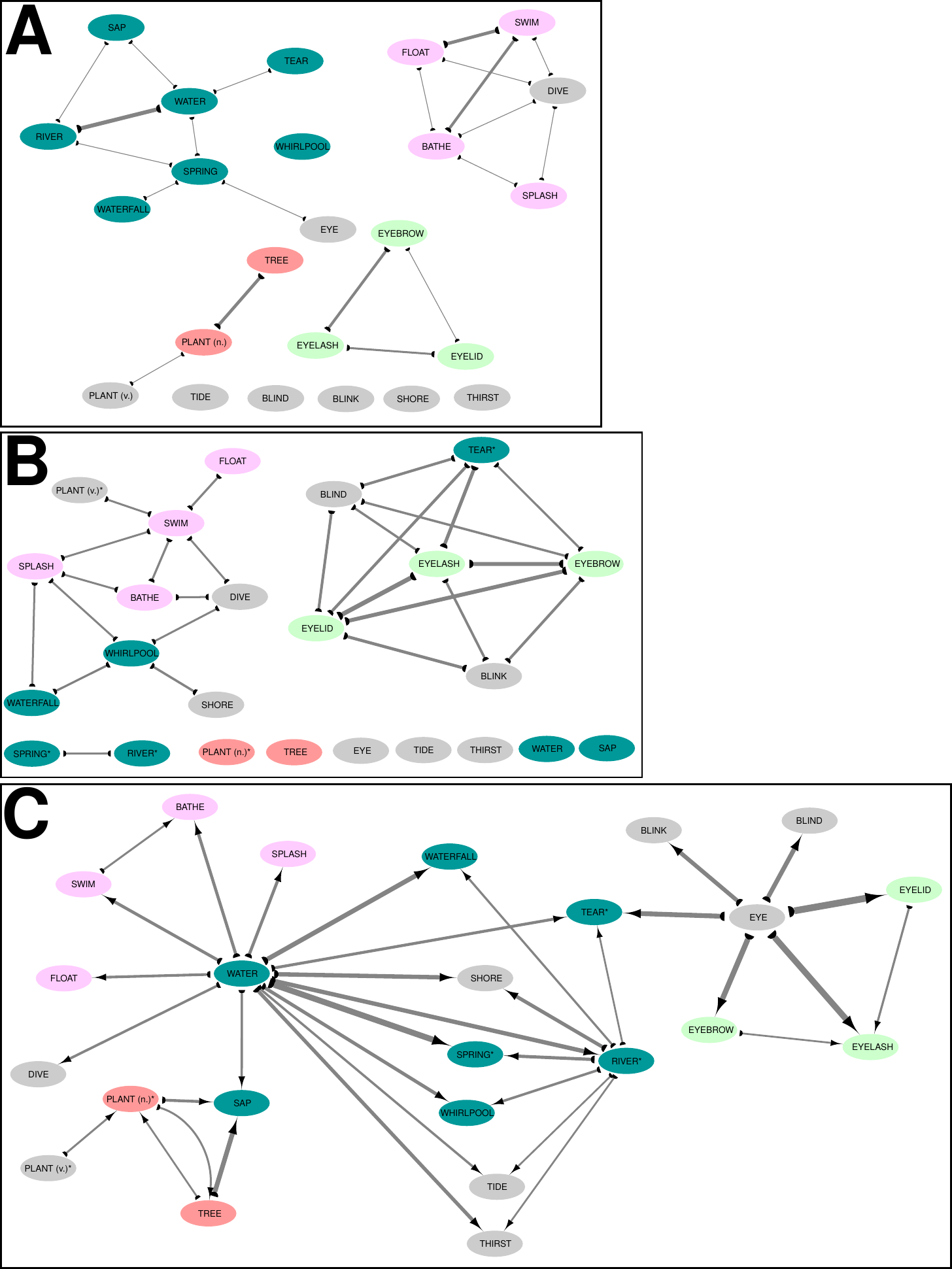}
    \caption{Comparing full (A), overlap (B), and affix (C) colexifications for subgraphs of the IDS dataset.
    Line width indicates the weight of the colexifications, colors other than light gray indicate
    communities inferred for the full colexification network, and link directions in the affix
    colexification network (B) are displayed with the help of arrows. Concept labels are taken from
    the Concepticon project. Concept labels with an asterisk were modified to
    to enhance the visualization.}
    \label{fig5}
\end{figure}

As can be seen from the visualizations shown in Figure \ref{fig5}, the three networks show a
remarkable difference in their individual structures, although they all involve the same concepts. 
Thus, while the concept \texttt{EYE} has only one spurious link in the full colexification network
to \texttt{SPRING} (A), it is completely isolated in the overlap colexification network (B), while
appearing as a rather central concept with a high out-degree in the affix colexification network
(C). 
When inspecting connected components in all three networks, we find huge differences between the
concepts that are fully connected with each other, while it is easy to spot semantic or
morphological connections that give rise to these patterns. Thus, 
we find a cluster of \texttt{BLIND}, \texttt{TEAR}, \texttt{EYELASH},
\texttt{EYEBROW}, \texttt{EYELID}, and \texttt{BLINK} in the overlap colexification network that
clearly seems to result from the fact that the words expressing these concepts all contain a
morpheme for \texttt{EYE}. The central position of \texttt{EYE} in the affix colexification network
confirms this role, and we find similar structures for \texttt{WATER} as another central concept in
the affix colexification network. 
A systematic comparison of these different kinds of colexification networks allows us to identify
semantic \emph{key players} that play an important role in contributing morphological material to
the construction of the lexicon of many of the world's languages.
\section{Discussion}
This study has presented new ideas regarding the inference of partial colexification networks from multilingual wordlists.  It has introduced new models that can be used to handle partial colexification patterns and proposed new efficient methods and workflows for the inference of partial colexification networks. Two new ways to handle partial colexification patterns in networks were introduced, namely \emph{affix colexifications} and \emph{overlap colexifications}. Using these new types of colexification patterns to infer affix and overlap colexification networks from large multilingual wordlist revealed some interesting properties of both network types. While overlap colexification networks allow us to measure the \emph{compoundhood} of individual concepts across the world's languages, affix colexification networks could be used as an initial proxy to measure \emph{lexical root productivity} across languages. Apart from being interesting for people working in the field of lexical typology, we assume that these new types of colexification networks can be very useful for many additional scientific fields in the future, including most notably computer science (and approaches to computational semantics) and psychology.

\section*{Funding}
This research was supported by the Max Planck Society Research Grant \textit{CALC³} (JML, \url{https://digling.org/calc/}), the ERC Consolidator Grant \textit{ProduSemy} (JML, Grant No. 101044282, see \url{ https://cordis.europa.eu/project/id/101044282}).
\section*{Acknowledgments}
I thank Robert Forkel for helpful comments on parts of the code base and the creation of the segmented version of the Intercontinental Dictionary Series.
\section*{Supplementary Materials}
The data and code used in this study are curated on GitHub, where they can be accessed at \url{https://github.com/lingpy/pacs/releases/tag/v0.1} (Version 0.1). Upon publication, data and code will also be archived with Zenodo.

\section*{Data Availability Statement}
Data and code accompanying the study are made freely available. 
The data and code used in this study are curated on GitHub, where they can be accessed at \url{https://github.com/lingpy/pacs/releases/tag/v0.1} (Version 0.1). Upon publication, data and code will also be archived with Zenodo.

\bibliographystyle{Frontiers-Harvard}

\bibliography{test}

\end{document}